\documentclass[conference]{IEEEtran}
\IEEEoverridecommandlockouts

\usepackage{cite}
\usepackage{amsmath,amssymb,amsfonts}
\usepackage{algorithmic}
\usepackage{graphicx}
\usepackage{textcomp}
\usepackage{xcolor}
\usepackage{booktabs}
\usepackage{multirow}
\usepackage{caption}
\usepackage{setspace}
\usepackage{url}
\usepackage{paralist}
\usepackage{marvosym}
\usepackage{xspace}
\usepackage{color}
\usepackage{multirow}
\usepackage{ifthen}

\renewcommand{\baselinestretch}{1}

\long\def\ignorethis#1{}

\definecolor{gray}{rgb}{0.35,0.35,0.35}
\definecolor{MyBlue}{rgb}{0,0.2,0.8}
\definecolor{MyRed}{rgb}{0.8,0.2,0}
\definecolor{MyGreen}{rgb}{0.0,0.5,0.1}
\definecolor{MyGray}{rgb}{0.4,0.4,0.4}

\newlength\paramargin
\newlength\figmargin
\newlength\subfigmargin
\newlength\secmargin
\newlength\subsecmargin
\newlength\tabmargin
\newlength\eqmargin

\usepackage{array}
\newcolumntype{L}[1]{>{\raggedright\let\newline\\\arraybackslash\hspace{0pt}}m{#1}}
\newcolumntype{C}[1]{>{\centering\let\newline\\\arraybackslash\hspace{0pt}}m{#1}}
\newcolumntype{R}[1]{>{\raggedleft\let\newline\\\arraybackslash\hspace{0pt}}m{#1}}

\def\eg{e.g.,~}

\newcommand{\secref}[1]{Section~\ref{sec:#1}}

\newcommand{\figref}[1]{Fig.~\ref{fig:#1}}
\newcommand{\tabref}[1]{Table~\ref{tab:#1}}

\newcommand{\Paragraph}[1]{\noindent\textbf{#1}}

\newcommand{\titlesupp}{
Camera Artist: A Multi-Agent Framework for Cinematic Language Storytelling Video Generation
}
\usepackage[ruled,vlined,linesnumbered]{algorithm2e}
\def\BibTeX{{\rm B\kern-.05em{\sc i\kern-.025em b}\kern-.08em
    T\kern-.1667em\lower.7ex\hbox{E}\kern-.125emX}}
\begin{document}

\def\maketitlesupplementary
{
   \newpage
   \twocolumn[
    \centering
    \Large
    \textbf{\titlesupp}\\
    \vspace{0.5em}Supplementary Material \\
    \vspace{1.0em}
   ]
}

\title{Camera Artist: A Multi-Agent Framework for Cinematic Language Storytelling Video Generation\\
\thanks{\textsuperscript{\Letter}Corresponding author}
}

\definecolor{mycolor_blue}{RGB}{231,239,250}
\definecolor{mycolor_green}{RGB}{230,247,224}
\definecolor{mycolor_gray}{RGB}{236,236,236}
\definecolor{pearDark!20}{RGB}{212,230,241}
\author{\IEEEauthorblockN{Haobo Hu\textsuperscript{1}, 
                          Qi Mao\textsuperscript{1, 2\Letter}, 
                          Yuanhang Li\textsuperscript{1}, 
                          and Libiao Jin\textsuperscript{1}}
\IEEEauthorblockA{\textsuperscript{1}School of Information and Communication Engineering, Communication University of China, Beijing, China}
\IEEEauthorblockA{\textsuperscript{2}State Key Laboratory for Multimedia Information Processing, \\ 
School of Computer Science, Peking University, Beijing, China}
\IEEEauthorblockA{Email: hhaobo@mails.cuc.edu.cn, \{qimao, yuanhangli, libiao\}@cuc.edu.cn}
}

\maketitle



\begin{abstract}
We propose \textbf{Camera Artist}, a multi-agent framework that models a real-world filmmaking workflow to generate narrative videos with explicit cinematic language. 
While recent multi-agent systems have made substantial progress in automating filmmaking workflows from scripts to videos, they often lack explicit mechanisms to structure narrative progression across adjacent shots and deliberate use of cinematic language, resulting in fragmented storytelling and limited filmic quality.
To address this, Camera Artist builds upon established agentic pipelines and introduces a dedicated \emph{Cinematography Shot Agent}, which integrates \textbf{recursive storyboard generation} to strengthen shot-to-shot narrative continuity and \textbf{cinematic language injection} to produce more expressive, film-oriented shot designs.
Extensive quantitative and qualitative results demonstrate that our approach consistently outperforms existing baselines in narrative consistency, dynamic expressiveness, and perceived film quality.

\end{abstract}

\begin{IEEEkeywords}
Narrative Video Generation, Multi-Agent Collaboration, Story Visualization , Cinematic Language, Image-to-video (I2V)
\end{IEEEkeywords}

\section{Introduction}
\label{sec:intro}

Film-making is a sophisticated art form where immersion and aesthetic impact derive not just from visual content, but from the deliberate design of cinematic language, \eg the precise orchestration of plot, camera movement, and lighting intended to guide emotion over time.
Inspired by this, creators seek to replicate film-level storytelling within AI-generated content (AIGC). 
Yet, despite the prowess of current Text-to-Video (T2V) and Image-to-Video (I2V) models~\cite{wan2025wanopenadvancedlargescale,wang2023modelscope,kong2024hunyuanvideo,li2024starvid,xiao2025captaincinemashortmovie} in producing high-fidelity short clips, they remain predominantly clip-centric, prioritizing local visual quality over the cinematic reasoning required to orchestrate multi-stage narratives.
Consequently, bridging the gap between visually striking fragments and coherent cinematic narratives remains a central challenge.

\begin{figure}[t]
    \centering
    \includegraphics[width=\linewidth]{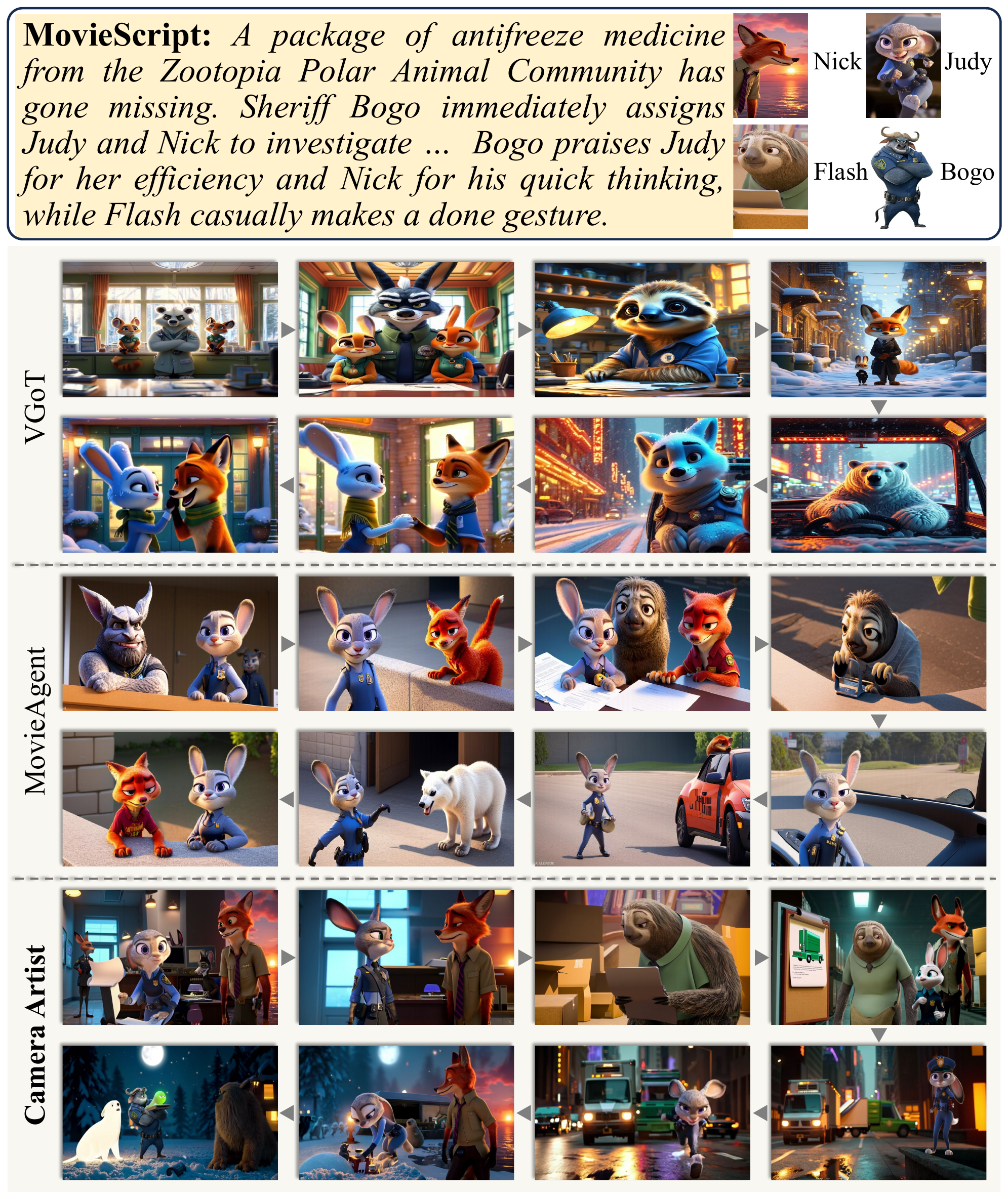}
    \vspace{-6 mm}
    \caption{
       \textbf{Comparison with multi-agent system on filmic storytelling.}
Existing multi-agent methods tend to exhibit fragmented narratives and weak cinematic control. In contrast, \emph{Camera Artist} achieves stronger shot-to-shot coherence and richer cinematic expression, yielding more filmic storytelling.
    }
    \label{fig:teaser}
\vspace{-7 mm}
\end{figure}

\begin{figure*}[!t]
    \centering
    \includegraphics[width=\linewidth]{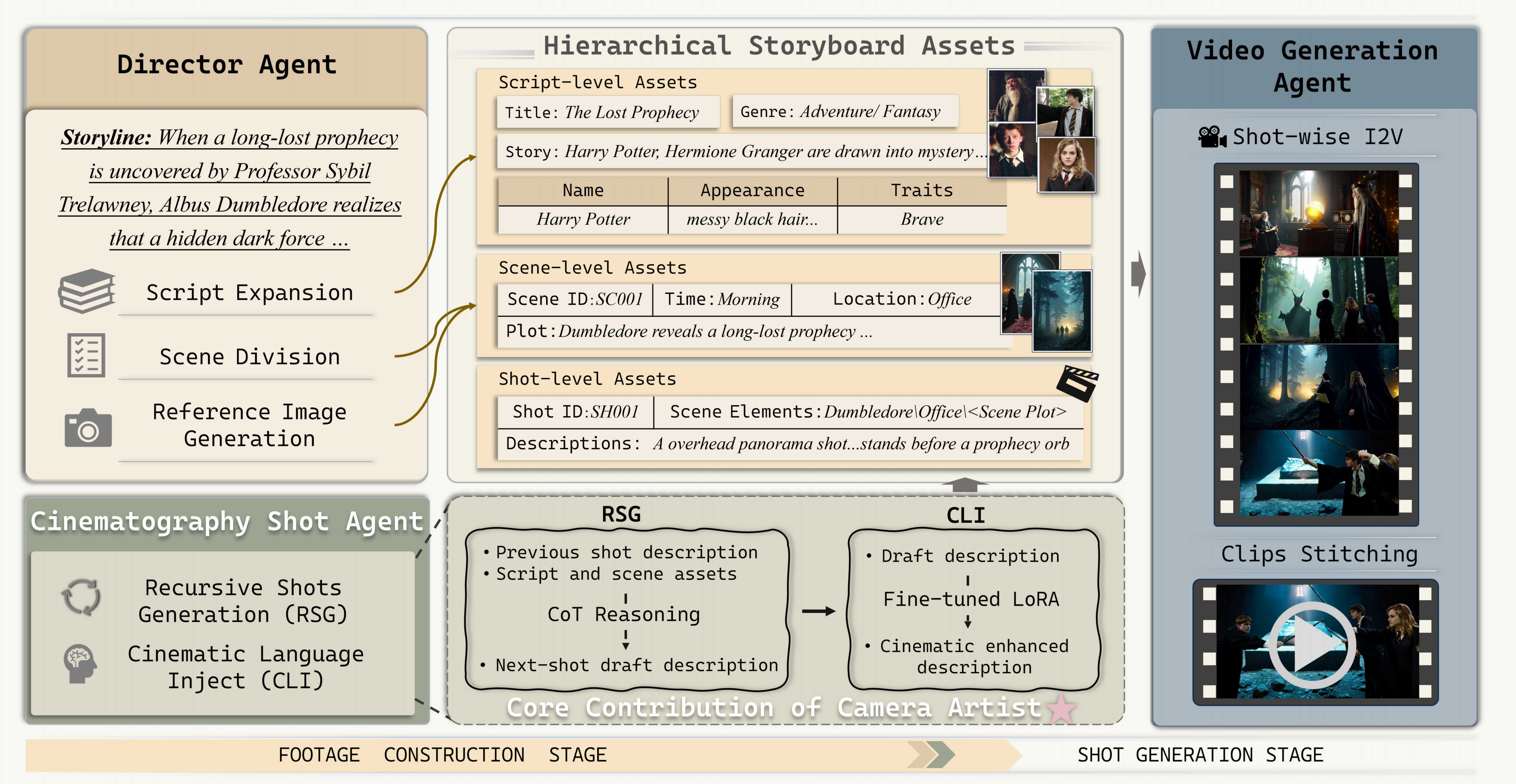}
    \vspace{-6mm}
    \caption{
       \textbf{The overall framework of Camera Artist.}
Camera Artist operates in two stages: \textit{footage construction} and \textit{shot generation}. 
In the footage construction stage, the Director Agent expands the story outline and builds hierarchical storyboard assets at script, scene, and shot levels. 
In the shot generation stage, the Cinematography Shot Agent first performs recursive shot generation to ensure narrative coherence, and then injects cinematic language to refine shot descriptions. 
Finally, the Video Generation Agent produces shot-wise videos and stitches them into a complete long-form narrative film.
    }
    \label{fig:framework}
\vspace{-7 mm}
\end{figure*}
To move beyond clip-level generation, multi-agent systems (MAS)~\cite{dorri2018multi} serve as a promising paradigm for long-form video production. 
By assigning Large Language Models (LLMs)~\cite{chang2024survey} to specialized roles—such as director, screenwriter, and cinematographer—these systems~\cite{10.1145/3680528.3687688,he2025dreamstoryopendomainstoryvisualization,zheng2024videogen,wu2025movieagent} mirror the collaborative workflow of professional film studios, which makes complex story generation feasible. 
However, as illustrated in \figref{teaser}, narrative consistency alone does not guarantee cinematic expressiveness. 
This discrepancy stems from the fact that existing MAS frameworks primarily focus on the logical alignment between scripts and visuals, often resulting in a mechanical assembly of scenes that lacks the deliberate authorship of a film. 
This limitation prompts a pivotal question: \emph{How can multi-agent video generation move beyond simple storytelling sequences to create videos that truly feel like cinema?}

The answer lies in two key limitations of existing frameworks.
First, current systems typically generate shot descriptions directly from scenes or scripts with limited conditioning on prior context, triggering ``narrative drift'' where adjacent shots fail to maintain fluid visual transitions.
Second, general-purpose LLMs acting as screenwriters often produce generic prompts rather than leveraging professional cinematic language to drive expressive visual storytelling.
These observations suggest that film-level generation requires both explicit modeling of narrative continuity and specialized cinematic injection.

\begin{figure}[t]
    \centering
    \includegraphics[width=1.\linewidth]{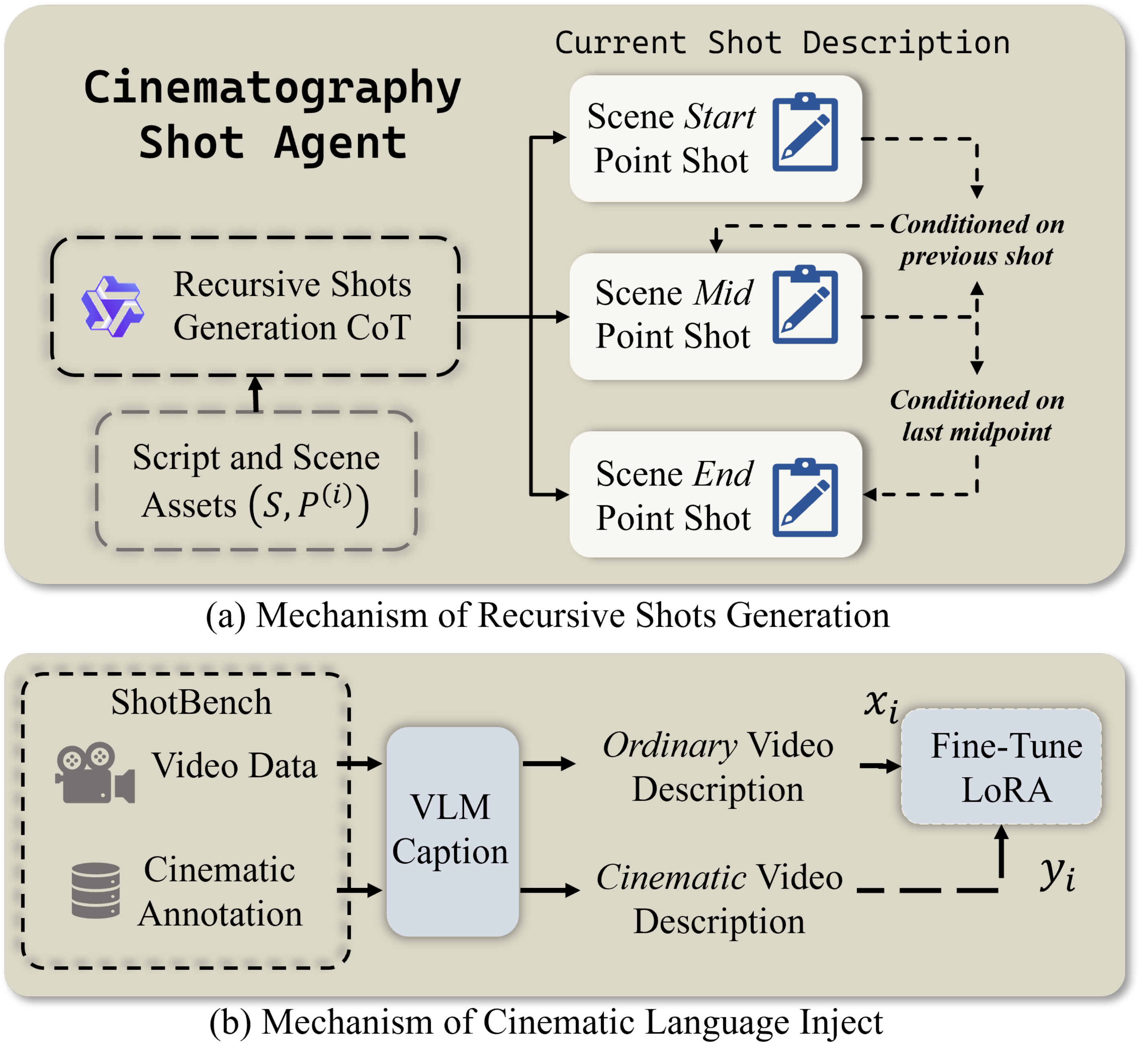} 
    \vspace{-8 mm}
        \caption{
            \textbf{Mechanism of the Cinematography Shot Agent.}
(a) Recursive Shots Generation (RSG): By recursively generating shots and selecting start/mid/end types, the system produces storyboards with strong narrative coherence.
(b) Cinematic Language Injection (CLI): A fine-tuned LLM trained on professional cinematic language transforms original shot descriptions into film-style, cinematically expressive ones.
        }
        \label{fig:Shot_Method}
\vspace{-8 mm}
\end{figure}

To address these challenges, we introduce \textbf{Camera Artist}, a multi-agent filmmaking framework designed for high-end cinematic storytelling. 
In our framework, the Director Agent oversees the narrative arc, while the Cinematography Shot Agent utilizes two novel mechanisms: \textbf{Recursive Shot Generation (RSG)} and \textbf{Cinematic Language Injection (CLI)}. 
Specifically, RSG enforces narrative continuity by conditioning each shot’s planning on the preceding shot’s context via a Chain-of-Thought (CoT)~\cite{wei2022chain} reasoning process, which ensures a logical and stylistic flow. 
Concurrently, CLI leverages a specialized LLM fine-tuned on a professional cinematography knowledge to translate abstract plot points into precise, film-oriented technical descriptions.
As demonstrated in \figref{teaser}, Camera Artist effectively strengthen the narrative continuity and cinematic expression across the production pipeline, resulting in a more cohesive and film-like storytelling experience.

Our main contributions are summarized as follows:
\begin{itemize}
    \item We introduce a multi-agent framework that automates the complete workflow of narrative video generation, from script understanding to cinematic shot planning and final rendering.
    \item We propose an explicit recursive shot generation module that enhances narrative coherence across shots, together with a cinematic language injection mechanism that enriches visual expression through purposeful shot language.
    \item Extensive experiments demonstrate that our method achieves superior narrative coherence, shot diversity, and temporal stability compared to existing baselines.
\end{itemize}

\section{Our Solution: Camera Artist}
\label{sec:Method}

In this section, we introduce \textbf{Camera Artist}, a multi-agent framework that transforms a user-provided story outline $O$ into a temporally ordered sequence of video clips $\mathcal{V}$.
Rather than rethinking the agentic paradigm, Camera Artist builds upon established multi-agent filmmaking workflows and targets two key factors of film-quality storytelling: shot-level narrative coherence and cinematic expressiveness.
We first present the overall workflow and agent roles in \secref{Director}, followed by the recursive shot generation and cinematic language injection modules in \secref{Recursive} and \secref{Cinematic}.
%

 \begin{figure*}[t]
    \centering
    \includegraphics[width=1.\textwidth]{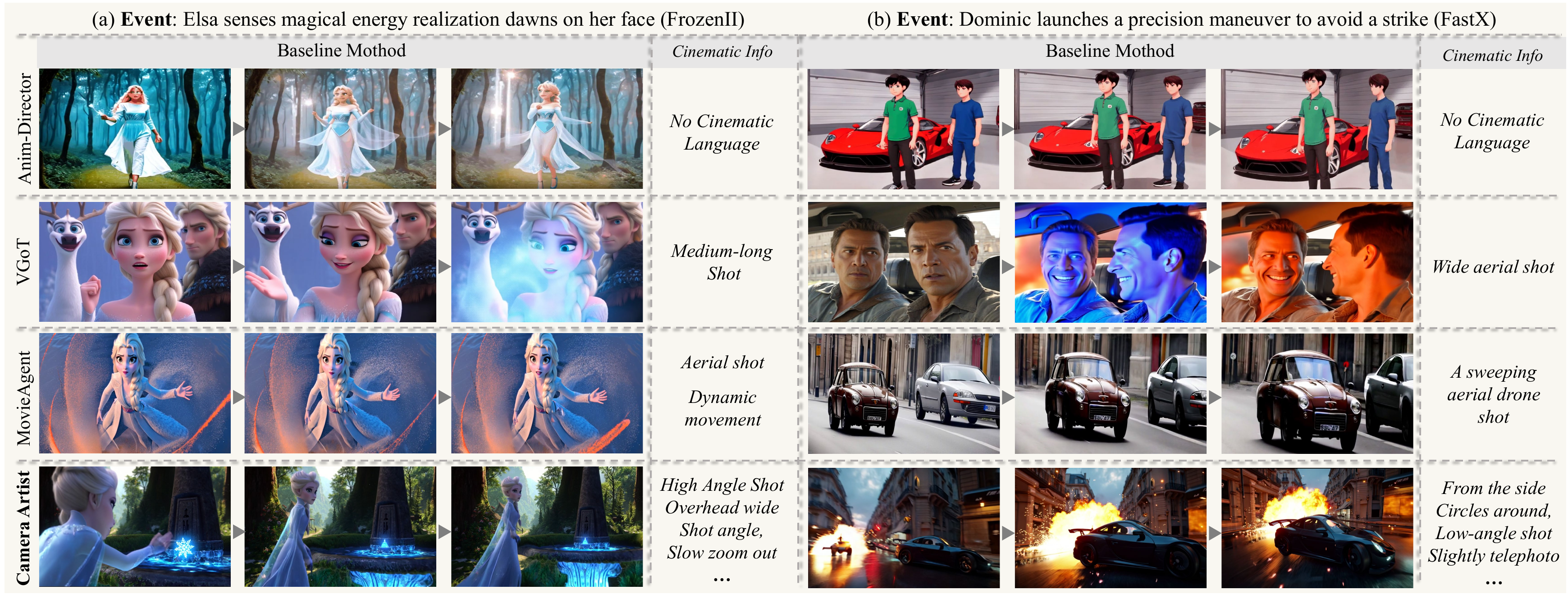} 
    \vspace{-7 mm}
    \caption{
\textbf{Qualitative experimental results of single shot content.}
For videos with similar shot content, Camera Artist can achieve richer and more expressive cinematic language, outperforming prior multi-agent methods.
    }
    \vspace{-6 mm}
    \label{fig:cammotion}
\end{figure*}

\vspace{-2 mm}
\subsection{Multi-Agent Collaborative System Framework}
\label{sec:Director}

As illustrated in \figref{framework}, Camera Artist consists of three collaborative agents: a \emph{Director Agent} for narrative planning, a \emph{Cinematography Shot Agent} for shot-level design with cinematic language, and a \emph{Video Generation Agent} for visual rendering.
The pipeline operates on a three-layer hierarchical storyboar and involves two stages: Footage Construction and Shot Generation.
In the \textbf{footage construction stage}, the Director Agent performs global narrative planning by decomposing the input story outline $O$ into script-level resources $S$, scene-level properties $\mathcal{P}$, and visual references $R$. 
Based on these resources, the Cinematography Shot Agent recursively generates an ordered sequence of shot descriptions $\mathcal{s}$ enriched with cinematic attributes.
These resources collectively constitute the storyboard representation $\mathcal{A}=\{S, P, \mathcal{s}\}$.
In the \textbf{shot generation stage}, the Video Generation Agent a multi-reference I2V model to generate video clips based on $\mathcal{s}$ and $R$.All video clips are concatenated to form the complete output video.
The overall workflow is detailed in the supplementary material \textbf{(SM)}. 




\Paragraph{Director Agent.}
The Director Agent serves as a global planner responsible for narrative expansion, scene decomposition, and visual reference construction.
Through structured CoT~\cite{wei2022chain} prompting, it expands the script-level narrative $S$ by refining genres, character identities, and storylines while strictly adhering to the original outline.
It further decomposes the script into an ordered sequence of scenes $\mathcal{P}=\{P^{(1)},\dots,P^{(k)}\}$, where each scene contains detailed information such as location, plot, and characters.
Additionally, based on character profiles and scene layouts, the Director Agent employs a T2I model to generate visual reference images $R$, which provide the foundation for subsequent shot generation and video rendering.


\Paragraph{Cinematography Shot Agent.}
Given each scene $P^{(j)}$ and the associated references $R$, the Cinematography Shot Agent recursively generates a sequence of shot descriptions enriched with cinematic language, ensuring both the cinematic expression of local shot clip and narrative coherence for global video. Each shot description $\mathcal{s}_j^i$ explicitly encodes action content, camera configuration, and visual composition.

\Paragraph{Video Generation Agent.}
The Video Generation Agent retrieves character- and scene-level references $R$ from $\mathcal{A}$ and conditions a multi-reference I2V model on both $R$ and the shot description $\mathcal{s}_j^i$ to generate a video clip $V_j^i$. 
This design can preserve identity consistency and spatial–temporal continuity across shots and scenes. All clips are finally concatenated to form the long-form narrative video.

\vspace{-1mm}
\subsection{Recursive Shots Generation}
\label{sec:Recursive}
\vspace{-2mm}
To enhance narrative coherence, we propose a Recursive Storyboard Generation method for the Cinematography Shot Agent.
Each shot is generated by conditioning on the global script and prior shots, simulating the human writing process of connecting sequential shots.
Given the $\{S,\mathcal{P}\}$ produced by the Director Agent, the Cinematography Shot Agent generates shots in scene order.
For each shot, the agent autonomously determines the shot content and type by conditioning on both scene and the prior shot information, and outputs the corresponding shot description.
As illustrated in \figref{Shot_Method}(a), we define shot types as follow: 
\begin{itemize}
\item \textbf{Scene Start Point $s^{1}_j$ :}
The first shot in the current scene $P^{(j)}$, directly generated without any previous shot description as an input, serving as the starting point for the recursive process.
\item \textbf{Scene Midpoint $s^{i}_j$ :} 
A common shot type that requires the previous shot content as a condition for its generation.
\item \textbf{Scene End Point $s^{N}_j$ :}The end point of the recursive shot generation process for the current scene $P^{(j)}$.
\end{itemize}

For scene $P^{(j)}$, shots are generated recursively:
\begin{equation} 
\vspace{-1 mm}
\mathrm{s}_{j, j \in \{1, \ldots, k\}.}^i = 
\begin{cases}
f\left( P^{(j)}, S \right), & i=1, \\
f\left( \mathrm{s}_j^{i-1}, P^{(j)}, S \right), & 2 \leq i \leq N .
\end{cases}
\vspace{-1 mm}
\end{equation}

When generating $\mathrm{s}_j^{i}$ for the $j$-th scene, the agent conditions on the scene $P^{(j)}$ and the previous shot $\mathrm{s}_j^{i-1}$ as contextual input, and the recursion stops once a $s^{N}_j$ is predicted.

\vspace{-2mm}
\subsection{Cinematic Language Injection}
\vspace{-1mm}
\label{sec:Cinematic}
To enhance film-level expressiveness in shot generation beyond narrative coherence, we introduce a \textbf{Cinematic Language Injection} mechanism for the Cinematography Shot Agent.
Built upon RSG, this module explicitly reasons about cinematic language by refining each shot with purposeful camera attributes, enabling the generated shots to better reflect professional cinematic language and visual intention.

We achieve cinematic language injection by fine-tuning a LLM with a Low-Rank Adaptation (LoRA) strategy~\cite{hu2021loralowrankadaptationlarge}. Specifically, as illustrated in \figref{Shot_Method}(b), 
we employ GPT-4o~\cite{gpt4o} to obtain an ordinary video description $x_i$ of the raw video, which focuses on objects and actions excluding cinematic cues.
%
We further utilize $x_i$ and shot-level cinematic annotations $d_i$ to generate a corresponding cinematic-enriched description $y_i$ with professional shot language descriptions via GPT-4o~\cite{gpt4o}. The mapping is formulated as follows:

\vspace{-3mm}
\begin{equation}
\vspace{-1 mm}
    y_i = f_{LLM}(x_i,d_i),
\vspace{-1mm}
\end{equation}
where, $f_{LLM} $ denotes the LLM mapping function. 
The optimization objective for LLM fine-tuning is formulated as follows:
\begin{equation}
    \mathcal{L}_{\text{cine}} = - \sum_{i=1}^{N} \log P_{\theta'}(y_i \,|\, x_i).
\vspace{-2 mm}
\end{equation}

%
During inference, the fine-tuned LLM injects explicit cinematic semantics into each recursively generated shot description, producing detailed scenes descriptions enriched with professional cinematic language.
%


\section{Experiments}
\label{sec:experiments}

\subsection{Experimental Setup}
\vspace{-1mm}
\Paragraph{Framework Configurations.}
We adopt Qwen3-30B-A3B-Instruct~\cite{yang2025qwen3technicalreport} as the default LLM backbone for all agents. 
We additionally fine-tune Qwen3-4B with LoRA~\cite{hu2021loralowrankadaptationlarge} for cinematic language injection using 580 curated paired samples $(x_i, y_i)$ from the ShotBench~\cite{liu2025shotbench} dataset.
The model is trained for 20 epochs with the Adam optimizer at a learning rate of $1\times10^{-4}$, applying LoRA with rank $8$ and scaling factor $32$ to all linear layers.
We employ MAGREF~\cite{deng2025magref}, which exhibits robust multi-reference controllability, as the video generator and utilize Flux \cite{labs2025flux1kontextflowmatching} to create high-quality reference images. 
All generated video clips feature a resolution of $832 \times 480$ at a frame rate of 15 fps.
All experiments are conducted on one NVIDIA Tesla A800 80G GPUs.

\Paragraph{Benchmark.}
We evaluate our framework on MoviePrompts ~\cite{wu2025movieagent}, which contains plot descriptions and character profiles from ten professional films.  
To further assess generalization, we construct an additional benchmark consisting of eight additional storytelling samples that follow the same format. 
\begin{table}[t]
\centering
\caption{\textbf{Quantitative comparison using VBench and CLIP-based semantic consistency.} Best and second-best results are highlighted in
\colorbox{pearDark!20}{blue} and \colorbox{mycolor_green}{green}, respectively.}
\vspace{-1mm}
\label{tab:quanl}
\renewcommand{\arraystretch}{0.95}
\resizebox{\linewidth}{!}{
\begin{tabular}{ccccccc}
\toprule
& \textbf{Semantic Consistency} & \multicolumn{5}{c}{\textbf{VBench Metrics}} \\
\cmidrule(r){2-2} \cmidrule(r){3-7}
\multicolumn{1}{c}{\multirow{-2}{*}{\textbf{Method/Metrics}}}  & \textbf{CLIP-T ($\uparrow$)} & \textbf{Subj.($\uparrow$)} & \textbf{Bg.($\uparrow$)} & \textbf{Motion($\uparrow$)} & \textbf{Dyn.($\uparrow$)} & \textbf{Aesth.($\uparrow$)} \\
\midrule
VGoT~\cite{zheng2024videogen} & \colorbox{mycolor_green}{28.15} & \colorbox{mycolor_green}{78.58} & \colorbox{pearDark!20}{97.93} & \colorbox{mycolor_green}{99.27} & 16.67 & \colorbox{mycolor_green}{68.73} \\
Anim-Director~\cite{10.1145/3680528.3687688}   & 23.86 & 67.79 & 94.15 & 96.54 & 39.78 & 67.24 \\
MovieAgent~\cite{wu2025movieagent} & 22.25 & 71.01 & 94.52 & 98.00 & \colorbox{mycolor_green}{76.27} & 65.63 \\
\textbf{Ours} & \colorbox{pearDark!20}{29.61} & \colorbox{pearDark!20}{79.54} & \colorbox{mycolor_green}{96.26} & \colorbox{pearDark!20}{99.32} & \colorbox{pearDark!20}{80.00} & \colorbox{pearDark!20}{69.51} \\
\bottomrule
\end{tabular}}
\vspace{-7 mm}
\end{table}
\Paragraph{Evaluation Metrics.}
Following MovieAgent~\cite{wu2025movieagent}, we further incorporate automated metrics from VBench~\cite{huang2023vbench} to assess video results across multiple dimensions, including Subject Consistency (Subj.), Background Consistency (Bg.), Motion Smoothness (Motion), Dynamic Degree (Dyn.), and Aesthetic Score (Aesth.). 

Additionally, we utilize CLIP-T~\cite{radford2021learning} for semantic consistency evaluation.
To move beyond traditional metrics and capture narrative coherence and cinematic expressiveness, we introduce a VLM-based automatic evaluation protocol, which is detailed in the \emph{SM}.
Given sampled video frames with corresponding descriptions, a VLM produces 1-5 scores for four criteria: \emph{Script Consistency}, \emph{Camera-Movement Consistency}, \emph{Video Quality}, and \emph{Real-Movie Similarity}. 
In our evaluation, we utilize GPT-4o~\cite{gpt4o}, Qwen3~\cite{yang2025qwen3technicalreport}, and Gemini-3~\cite{gemini2023} as evaluators to provide a multifaceted measurement and mitigate potential biases inherent in any single model.

\Paragraph{Compared Methods.}
To evaluate the effectiveness of Camera Artist, we compare it with recent multi-agent video-generation systems, including VideoGen-of-Thought (VGoT)~\cite{zheng2024videogen}, Anim-Director~\cite{10.1145/3680528.3687688}, and MovieAgent~\cite{wu2025movieagent}.

\begin{table*}[!t]
\centering
\caption{
\textbf{Multi-VLM evaluation across narrative and cinematic dimensions.}
Best and second-best results are highlighted in
\colorbox{pearDark!20}{blue} and \colorbox{mycolor_green}{green}, respectively.
}
\vspace{-2 mm}
\label{tab:llm_eval_multi}

\renewcommand{\arraystretch}{1.15}
\setlength{\tabcolsep}{4pt}

\resizebox{\linewidth}{!}{%
\begin{tabular}{ccccccccccccccccc}
\toprule
& \multicolumn{4}{c}{\textbf{Script Cons.}($\uparrow$)}
& \multicolumn{4}{c}{\textbf{Cam. Cons.}($\uparrow$)}
& \multicolumn{4}{c}{\textbf{Video Qual.}($\uparrow$)}
& \multicolumn{4}{c}{\textbf{Real. Sim.}($\uparrow$)} \\
\cmidrule(r){2-5} \cmidrule(r){6-9} \cmidrule(r){10-13} \cmidrule(r){14-17}

\multicolumn{1}{c}{\multirow{-2}{*}{\textbf{Method/Metrics}}}
& GPT-4o & Qwen3 & Gemini & Avg.
& GPT-4o & Qwen3 & Gemini & Avg.
& GPT-4o & Qwen3 & Gemini & Avg.
& GPT-4o & Qwen3 & Gemini & Avg. \\
\midrule

VGoT~\cite{zheng2024videogen}
& 3.33 & 3.00 & 2.17 & 2.83
& 1.83 & 1.00 & 1.17 & 1.33
& \colorbox{mycolor_green}{4.67} & \colorbox{pearDark!20}{4.83} & 2.67 & 4.06
& \colorbox{mycolor_green}{4.17} & 3.67 & 2.67 & 3.50 \\

Anim-Director~\cite{10.1145/3680528.3687688}
& \colorbox{mycolor_green}{3.60} & 2.50 & 2.83 & \colorbox{mycolor_green}{2.98}
& \colorbox{mycolor_green}{2.40} & 1.00 & \colorbox{pearDark!20}{3.50} & 2.30
& 3.33 & 3.17 & 2.67 & 3.06
& 2.00 & 1.83 & 1.00 & 1.61 \\

MovieAgent~\cite{wu2025movieagent}
& 2.10 & 1.30 & \colorbox{mycolor_green}{3.17} & 2.19
& 1.70 & 1.00 & \colorbox{mycolor_green}{3.38} & \colorbox{mycolor_green}{2.03}
& 4.10 & 4.30 & \colorbox{mycolor_green}{4.00} & \colorbox{mycolor_green}{4.13}
& 2.90 & 2.70 & \colorbox{mycolor_green}{3.80} & \colorbox{mycolor_green}{3.13} \\

\textbf{Ours}
& \colorbox{pearDark!20}{4.50} & \colorbox{pearDark!20}{4.00} & \colorbox{pearDark!20}{3.20} & \colorbox{pearDark!20}{3.90}
& \colorbox{pearDark!20}{3.25} & \colorbox{pearDark!20}{3.90} & \colorbox{pearDark!20}{3.50} & \colorbox{pearDark!20}{3.55}
& \colorbox{pearDark!20}{4.86} & \colorbox{mycolor_green}{4.78} & \colorbox{pearDark!20}{4.50} & \colorbox{pearDark!20}{4.71}
& \colorbox{pearDark!20}{4.00} & \colorbox{pearDark!20}{4.56} & \colorbox{pearDark!20}{3.50} & \colorbox{pearDark!20}{4.02} \\

\bottomrule
\end{tabular}
}
\vspace{-3 mm}
\end{table*}

\begin{table*}[!t]
\centering
\caption{
\textbf{Quantitative results of the ablation study on recursive storyboard generation and cinematic language injection.}
Best and second-best results are highlighted in
\colorbox{pearDark!20}{blue} and \colorbox{mycolor_green}{green}, respectively.
}
\vspace{-2 mm}
\label{tab:unified_ablation}

\renewcommand{\arraystretch}{1.15}
\setlength{\tabcolsep}{4pt}

\resizebox{\linewidth}{!}{%
\begin{tabular}{ccccccccccc}
\toprule
& \textbf{Semantic} 
& \multicolumn{5}{c}{\textbf{VBench Metrics}} 
& \multicolumn{4}{c}{\textbf{LLM-based Evaluation}} \\

\cmidrule(r){2-2}
\cmidrule(r){3-7}
\cmidrule(r){8-11}

\multicolumn{1}{c}{\multirow{-2}{*}{\textbf{Method}}}
& \textbf{CLIP-T}
& \textbf{Subj.($\uparrow$)}
& \textbf{Bg.($\uparrow$)}
& \textbf{Motion($\uparrow$)}
& \textbf{Dyn.($\uparrow$)}
& \textbf{Aesth.($\uparrow$)}
& \textbf{Script Cons.($\uparrow$)}
& \textbf{Cam. Cons.($\uparrow$)}
& \textbf{Video Qual.($\uparrow$)}
& \textbf{Real. Sim.($\uparrow$)} \\
\midrule

w/o RSG
& 28.22
& \colorbox{mycolor_green}{74.69}
& 93.93
& \colorbox{mycolor_green}{99.04}
& \colorbox{mycolor_green}{78.67}
& \colorbox{mycolor_green}{67.45}
& 3.55
& \colorbox{mycolor_green}{3.36}
& \colorbox{mycolor_green}{4.45}
& \colorbox{mycolor_green}{3.91} \\

w/o CLI
& \colorbox{mycolor_green}{29.27}
& 73.49
& \colorbox{mycolor_green}{94.33}
& 98.97
& 74.25
& 67.10
& \colorbox{mycolor_green}{3.60}
& 2.83
& 4.00
& 3.67 \\

\textbf{Camera Artist}
& \colorbox{pearDark!20}{29.61}
& \colorbox{pearDark!20}{79.54}
& \colorbox{pearDark!20}{96.26}
& \colorbox{pearDark!20}{99.32}
& \colorbox{pearDark!20}{80.00}
& \colorbox{pearDark!20}{69.51}
& \colorbox{pearDark!20}{3.90}
& \colorbox{pearDark!20}{3.55}
& \colorbox{pearDark!20}{4.71}
& \colorbox{pearDark!20}{4.02} \\

\bottomrule
\end{tabular}
}
\vspace{-7 mm}
\end{table*}


\vspace{-3 mm}
\subsection{Comparison with Baseline}
\Paragraph{Quantitative Results.}
As shown in \tabref{quanl} and \tabref{llm_eval_multi}, our Camera Artist exhibits superior performance across all evaluated metrics. 
while VGoT~\cite{zheng2024videogen} reports the highest subject consistency, this is primarily attributed to its tendency to generate near-static videos, as evidenced by its lowest scores in dynamic degree. 
In contrast, our method achieves the highest motion dynamics while simultaneously maintaining high background consistency. 
Furthermore, the VLM-based evaluation in \tabref{llm_eval_multi} corroborates this trend; all evaluators indicate that Camera Artist performs exceptionally well in narrative coherence, camera movement, video quality, and cinematic realism.

\begin{figure}[t]
    \centering
    \includegraphics[width=1.\linewidth]{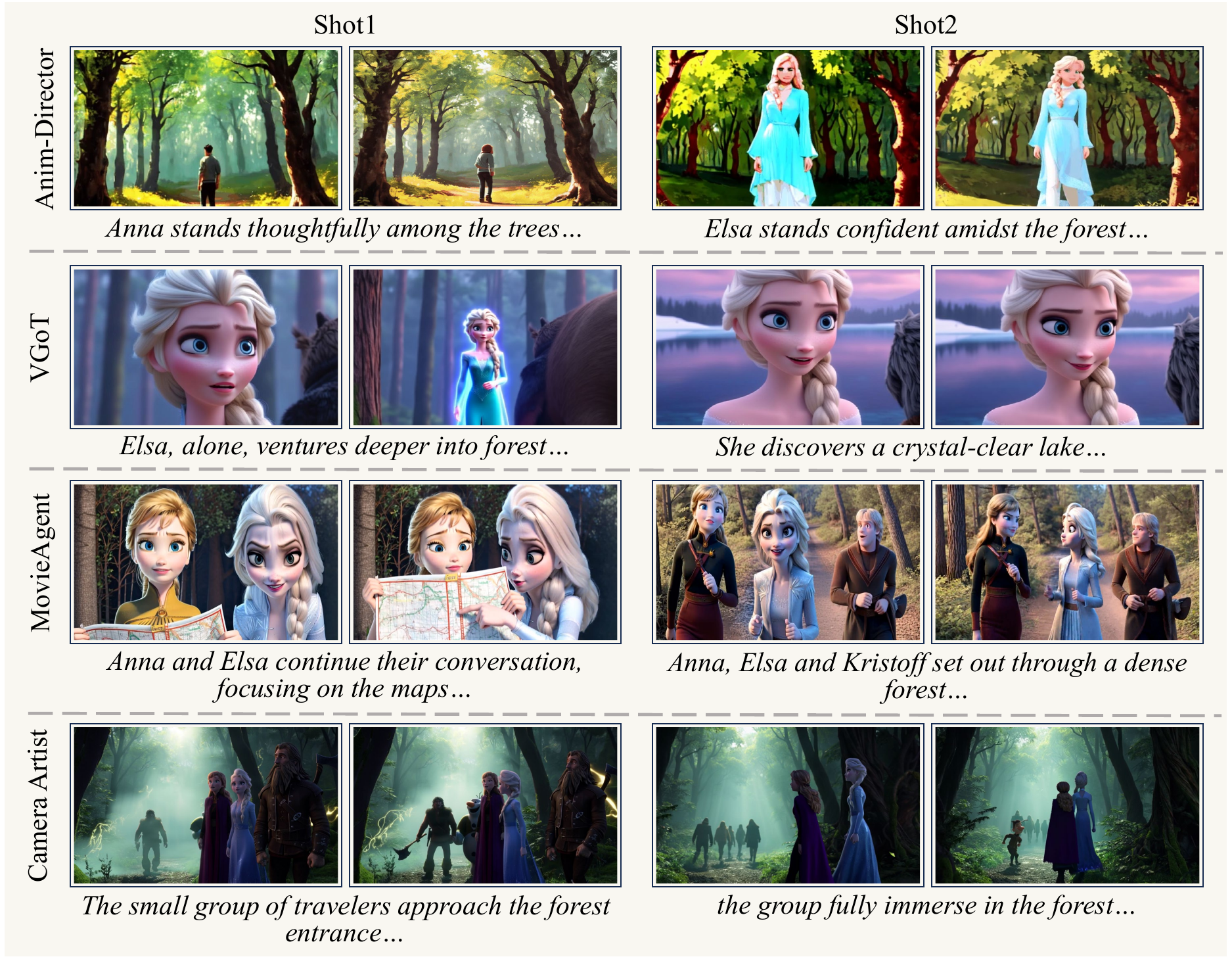} 
    \vspace{-6 mm}
\caption{\textbf{Qualitative comparison of inter-shot narrative coherence.}
Camera Artist conditions each shot on preceding shot and scene information, producing shot content that is narratively coherent in both text and visual realization.}
\vspace{-8 mm}
    \label{fig:diversity1}
\end{figure}


\Paragraph{Qualitative Results.}
%
\figref{cammotion} and \figref{diversity1} illustrate the qualitative advantages of Camera Artist in both single-shot cinematic expressiveness and multi-shot narrative coherence.
In single-shot scenarios, baseline methods often lack explicit cinematic guidance or rely on coarse camera specifications, leading to static or weakly expressive visuals.
For example, when the prompt specifies “Elsa senses magical energy,” Anim-Director~\cite{10.1145/3680528.3687688} produces visually similar shots, VGoT~\cite{zheng2024videogen} yields a fixed mid-to-long shot, and MovieAgent~\cite{wu2025movieagent} generates a largely static close-up.
In contrast, Camera Artist adopts ``a high-angle wide shot with a smooth zoom-out'', expanding spatial perception and strengthening cinematic impact.

Furthermore, baseline methods struggle to maintain narrative and visual continuity across adjacent shots.
In the example where “Elsa and Anna’s group ventures into  forest and ancient artifacts,” Anim-Director~\cite{10.1145/3680528.3687688} exhibits abrupt protagonist switching from ``Anna to Elsa'', resulting in fragmented storytelling with little visual or narrative linkage.
While VGoT~\cite{zheng2024videogen} and MovieAgent~\cite{wu2025movieagent} maintain better textual continuity at the shot level, yet their generated videos suffer from scene inconsistency: VGoT~\cite{zheng2024videogen} abruptly shifts from ``a forest'' to ``a lakeside'' and MovieAgent~\cite{wu2025movieagent} transitions from ``a nighttime forest'' to ``a daytime woodland path'', which breaks temporal and spatial coherence.
In contrast, Camera Artist preserves both character and scene consistency, coherently portraying the group’s progression from initial entry into the forest to deeper exploration, yielding a continuous narrative flow.

\begin{figure}[!t]
    \centering
    \includegraphics[width=\linewidth]{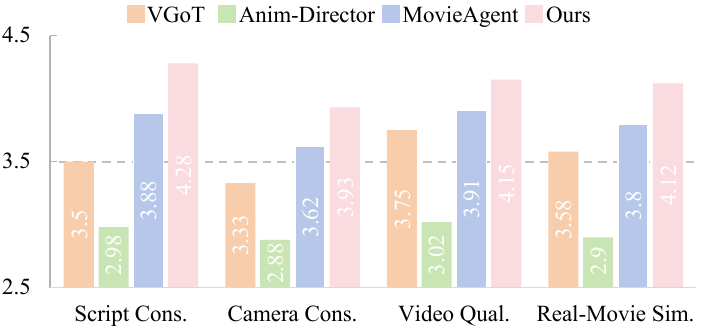}
    \vspace{-6 mm}
\caption{\textbf{User study comparison on four subjective metrics.}
Results of VGoT~\cite{zheng2024videogen}, Anim-Director~\cite{10.1145/3680528.3687688}, MovieAgent~\cite{wu2025movieagent}, and our method on Script Consistency, Camera-Movement Consistency, Video Quality, and Real-Movie Similarity. Our method achieves the highest scores across all metrics.}
    \label{fig:user}
\vspace{-8 mm}
\end{figure}

\vspace{-3.5 mm}
\subsection{User Study}
\label{sec:Userstudy}
\vspace{-1 mm}
Given the inherent subjectivity in cinematic quality and narrative perception, we conduct a human evaluation using a five-point Likert scale. 
This study assesses four key dimensions: \emph{Script Consistency}, \emph{Camera-Movement Consistency}, \emph{Video Quality}, and \emph{Real-Movie Similarity}.
During the evaluation, each participant is presented with the input script alongside video sequences generated by our method and the baselines. 
These sequences are displayed in a randomized order to mitigate potential ordering bias. 
As illustrated in \figref{user}, \textbf{Camera Artist} consistently achieves the highest aggregate scores across all evaluation dimensions. 
Specifically, our method reaches $4.28$ in script consistency and $4.12$ in Real-Movie Similarity, significantly outperforming the baselines. 
These results demonstrate that videos produced by Camera Artist are perceived as more coherent and cinematically compelling by human evaluators.

\begin{figure}[t]
    \centering
    \includegraphics[width=\linewidth]{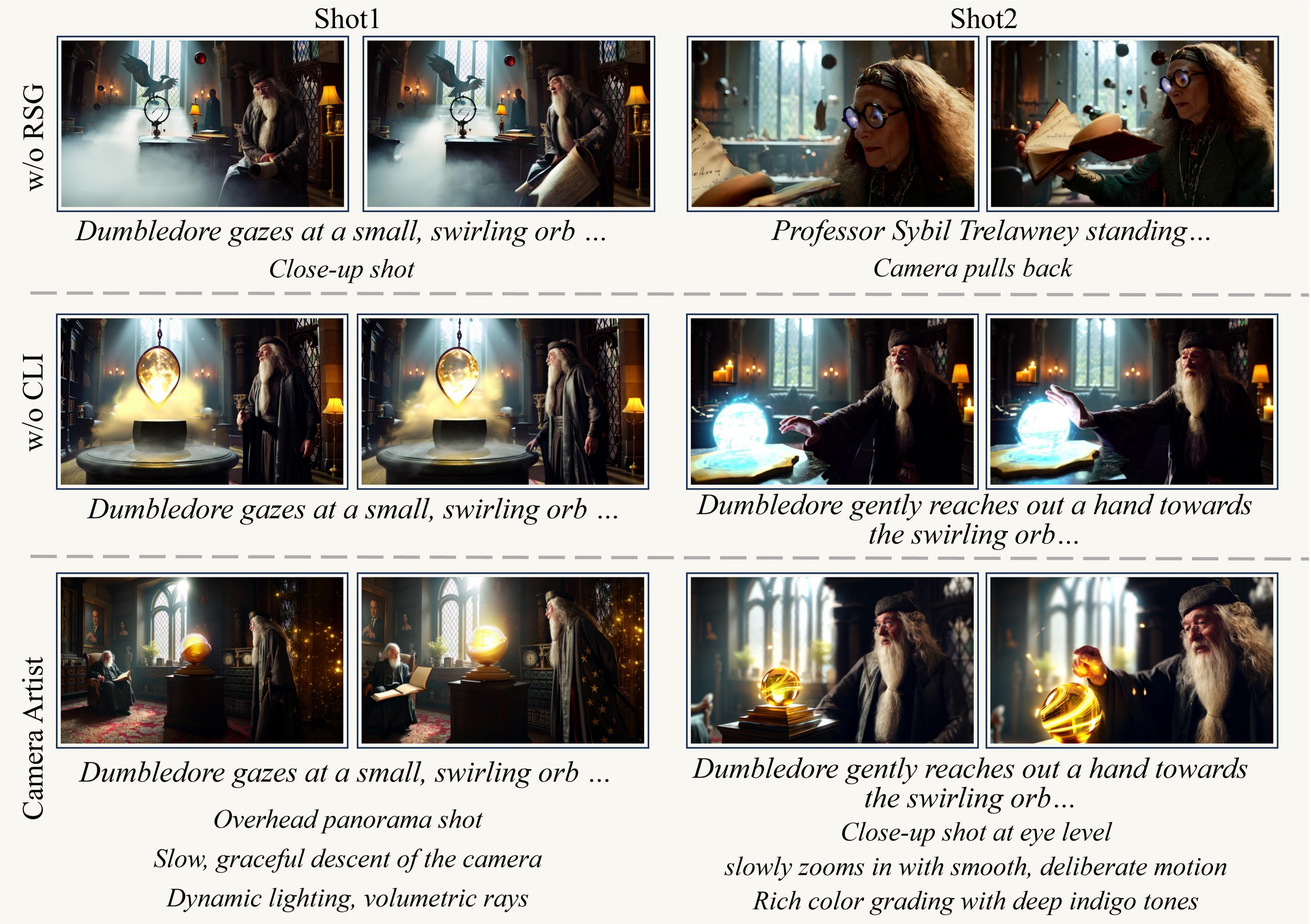}
    \vspace{-5 mm}
    \caption{
       \textbf{Ablation study on RSG and CLI.}
RSG preserves coherent shot-to-shot narrative flow, while CLI enhances cinematic expressiveness through deliberate camera motion and lighting; removing either results in fragmented storytelling or visually static shots.
    }
    \label{fig:abla}
\vspace{-7 mm}
\end{figure}

\vspace{-2mm}
\subsection{Ablation Study}
\vspace{-2 mm}
To evaluate the contribution of the core modules, we conduct an ablation study on \textit{(i)} \emph{RSG} and \textit{(ii)} \emph{CLI}. 
Quantitative and qualitative results are presented in \tabref{unified_ablation} and \figref{abla}, respectively.
As illustrated in \figref{abla}, the removal of RSG significantly diminishes narrative coherence across shots.
This leads to abrupt protagonist shifts, such as a sudden transition to a new character in the second shot, which disrupts the logical continuity and narrative rhythm.
This degradation is further evidenced by the script consistency scores in \tabref{unified_ablation}, where the configuration without RSG yields the lowest performance.
Furthermore, the exclusion of CLI results in a substantial decline in camera motion fidelity, with the score dropping from $3.55$ to $2.83$. 
In this case, the generated videos remain largely static and purely descriptive, failing to execute dynamic camera maneuvers.
In contrast, the full Camera Artist model, which integrates both RSG and CLI, produces a seamless narrative with deliberate camera motion, angles, and lighting that enhance the overall cinematic quality.

\section{Conclusions}
In this work, we propose \emph{Camera Artist}, a multi-agent framework for cinematic language storytelling video generation.
By integrating recursive storyboard generation and explicit cinematic language injection into an automated filmmaking pipeline, Camera Artist improves narrative coherence and film-level visual expressiveness beyond conventional clip-centric generation.
Extensive evaluations demonstrate the superior performance of our approach in both storytelling consistency and cinematic quality.
Overall, Camera Artist provides a robust framework for cinematic narrative generation, advancing the development of fully automated, professional-grade cinematic production systems.

\bibliographystyle{IEEEbib}
\bibliography{main}

@String(PAMI = {IEEE Trans. Pattern Anal. Mach. Intell.})

@String(CVPR= {IEEE Conf. Comput. Vis. Pattern Recog.})

@String(ICLR = {Int. Conf. Learn. Represent.})

@String(PAMI  = {IEEE TPAMI})

@String(CVPR  = {CVPR})

@String(TIST  = {ACM TIST})

@String(ICLR  = {ICLR})

@article{deng2025magref,
  title={MAGREF: Masked Guidance for Any-Reference Video Generation},
  author={Deng, Yufan and Guo, Xun and Yin, Yuanyang and Fang, Jacob Zhiyuan and Yang, Yiding and Wang, Yizhi and Yuan, Shenghai and Wang, Angtian and Liu, Bo and Huang, Haibin and others},
  journal={arXiv preprint arXiv:2505.23742},
  year={2025}
}

@article{kong2024hunyuanvideo,
  title={Hunyuanvideo: A systematic framework for large video generative models},
  author={Kong, Weijie and Tian, Qi and Zhang, Zijian and Min, Rox and Dai, Zuozhuo and Zhou, Jin and Xiong, Jiangfeng and Li, Xin and Wu, Bo and Zhang, Jianwei and others},
  journal={arXiv preprint arXiv:2412.03603},
  year={2024}
}

@article{wan2025wanopenadvancedlargescale,
  title={Wan: Open and advanced large-scale video generative models},
  author={Wan, Team and Wang, Ang and Ai, Baole and Wen, Bin and Mao, Chaojie and Xie, Chen-Wei and Chen, Di and Yu, Feiwu and Zhao, Haiming and Yang, Jianxiao and others},
  journal={arXiv preprint arXiv:2503.20314},
  year={2025}
}

@article{xiao2025captaincinemashortmovie,
  title={Captain cinema: Towards short movie generation},
  author={Xiao, Junfei and Yang, Ceyuan and Zhang, Lvmin and Cai, Shengqu and Zhao, Yang and Guo, Yuwei and Wetzstein, Gordon and Agrawala, Maneesh and Yuille, Alan and Jiang, Lu},
  journal={arXiv preprint arXiv:2507.18634},
  year={2025}
}

@article{wu2025movieagent,
  title={Automated movie generation via multi-agent cot planning},
  author={Wu, Weijia and Zhu, Zeyu and Shou, Mike Zheng},
  journal={arXiv preprint arXiv:2503.07314},
  year={2025}
}

@inproceedings{radford2021learning,
  title={Learning transferable visual models from natural language supervision},
  author={Radford, Alec and Kim, Jong Wook and Hallacy, Chris and Ramesh, Aditya and Goh, Gabriel and Agarwal, Sandhini and Sastry, Girish and Askell, Amanda and Mishkin, Pamela and Clark, Jack and others},
  booktitle={ICML},
  pages={8748--8763},
  year={2021},
}

@article{chang2024survey,
  title={A survey on evaluation of large language models},
  author={Chang, Yupeng and Wang, Xu and Wang, Jindong and Wu, Yuan and Yang, Linyi and Zhu, Kaijie and Chen, Hao and Yi, Xiaoyuan and Wang, Cunxiang and Wang, Yidong and others},
  journal={TIST},
  year={2024},
}

@article{dorri2018multi,
  title={Multi-agent systems: A survey},
  author={Dorri, Ali and Kanhere, Salil S and Jurdak, Raja},
  journal={IEEE Access},
  year={2018},
}

@article{he2025dreamstoryopendomainstoryvisualization,
      title={Dreamstory: Open-domain story visualization by llm-guided multi-subject consistent diffusion},
      author={He, Huiguo and Yang, Huan and Tuo, Zixi and Zhou, Yuan and Wang, Qiuyue and Zhang, Yuhang and Liu, Zeyu and Huang, Wenhao and Chao, Hongyang and Yin, Jian},
      journal={PAMI},
      year={2025},
      publisher={IEEE}
}

@article{yang2025qwen3technicalreport,
  title={Qwen3 technical report},
  author={Yang, An and Li, Anfeng and Yang, Baosong and Zhang, Beichen and Hui, Binyuan and Zheng, Bo and Yu, Bowen and Gao, Chang and Huang, Chengen and Lv, Chenxu and others},
  journal={arXiv preprint arXiv:2505.09388},
  year={2025}
}

@article{liu2025shotbench,
  title={ShotBench: Expert-Level Cinematic Understanding in Vision-Language Models},
  author={Liu, Hongbo and He, Jingwen and Jin, Yi and Zheng, Dian and Dong, Yuhao and Zhang, Fan and Huang, Ziqi and He, Yinan and Li, Yangguang and Chen, Weichao and others},
  journal={arXiv preprint arXiv:2506.21356},
  year={2025}
}

@article{labs2025flux1kontextflowmatching,
  title={FLUX. 1 Kontext: Flow Matching for In-Context Image Generation and Editing in Latent Space},
  author={Batifol, Stephen and Blattmann, Andreas and Boesel, Frederic and Consul, Saksham and Diagne, Cyril and Dockhorn, Tim and English, Jack and English, Zion and Esser, Patrick and others},
  journal={arXiv e-prints},
  pages={arXiv--2506},
  year={2025}
}

@inproceedings{10.1145/3680528.3687688,
author = {Li, Yunxin and Shi, Haoyuan and Hu, Baotian and Wang, Longyue and Zhu, Jiashun and Xu, Jinyi and Zhao, Zhen and Zhang, Min},
title = {Anim-Director: A Large Multimodal Model Powered Agent for Controllable Animation Video Generation},
year = {2024},
url = {https://doi.org/10.1145/3680528.3687688},
booktitle = {SIGGRAPH Asia},
}

@InProceedings{huang2023vbench,
      title={{VBench}: Comprehensive Benchmark Suite for Video Generative Models},
      author={Huang, Ziqi and He, Yinan and Yu, Jiashuo and Zhang, Fan and Si, Chenyang and Jiang, Yuming and Zhang, Yuanhan and Wu, Tianxing and Jin, Qingyang and Chanpaisit, Nattapol and Wang, Yaohui and Chen, Xinyuan and Wang, Limin and others},
      booktitle={CVPR},
      year={2024}
}

@inproceedings{wei2022chain,
  title={Chain-of-thought prompting elicits reasoning in large language models},
  author={Wei, Jason and Wang, Xuezhi and Schuurmans, Dale and Bosma, Maarten and Xia, Fei and Chi, Ed and Le, Quoc V and Zhou, Denny and others},
  booktitle={NeurIPS},
  year={2022}
}

@article{zheng2024videogen,
  title={VideoGen-of-Thought: Step-by-step generating multi-shot video with minimal manual intervention},
  author={Zheng, Mingzhe and Xu, Yongqi and Huang, Haojian and Ma, Xuran and Liu, Yexin and Shu, Wenjie and Pang, Yatian and Tang, Feilong and others},
  journal={arXiv preprint arXiv:2412.02259},
  year={2024}
}

@misc{gpt4o,
  title  = {GPT-4o},
  howpublished  = {Accessed May 13, 2024 [Online] \url{https://openai.com/index/hello-gpt-4o/}},
  url  = "https://openai.com/index/hello-gpt-4o/"
}

@techreport{gemini2023,
  title={Gemini: a family of highly capable multimodal models},
  author={Team, Gemini and Anil, Rohan and Borgeaud, Sebastian and Alayrac, Jean-Baptiste and Yu, Jiahui and Soricut, Radu and Schalkwyk, Johan and Dai, Andrew M and Hauth, Anja and Millican, Katie and others},
  journal={arXiv preprint arXiv:2312.11805},
  year={2023}
}

@inproceedings{hu2021loralowrankadaptationlarge,
      title={Lora: Low-rank adaptation of large language models.},
      author={Hu, Edward J and Shen, Yelong and Wallis, Phillip and Allen-Zhu, Zeyuan and Li, Yuanzhi and Wang, Shean and Wang, Lu and Chen, Weizhu and others},
      booktitle={ICLR},
      year={2022}
}

@article{li2024starvid,
  title={StarVid: Enhancing Semantic Alignment in Video Diffusion Models via Spatial and SynTactic Guided Attention Refocusing},
  author={Li, Yuanhang and Mao, Qi and Chen, Lan and Fang, Zhen and Tian, Lei and Xiao, Xinyan and Jin, Libiao and Wu, Hua},
  journal={arXiv preprint arXiv:2409.15259},
  year={2024}
}

@article{wang2023modelscope,
  title={Modelscope text-to-video technical report},
  author={Wang, Jiuniu and Yuan, Hangjie and Chen, Dayou and Zhang, Yingya and Wang, Xiang and Zhang, Shiwei},
  journal={arXiv preprint arXiv:2308.06571},
  year={2023}
}
\newpage

\setcounter{page}{1}
\maketitlesupplementary
\appendix

In this supplementary material, we present additional more implementation details and additional results as follows:

\begin{itemize}
\item In \secref{imple_details}, we provide additional implementation details on the progress of Camera Artist, including metrics of VLM-based evaluation, user studies, baselines, and quantitative metrics.
\item In \secref{Additional}, we also present additional qualitative results to enhance this paper.
\end{itemize}

\section{Implementation Details}
\label{sec:imple_details}

\subsection{Workflow Overview}
\begin{figure*}[!t]
    \centering
    \includegraphics[width=1.\linewidth]{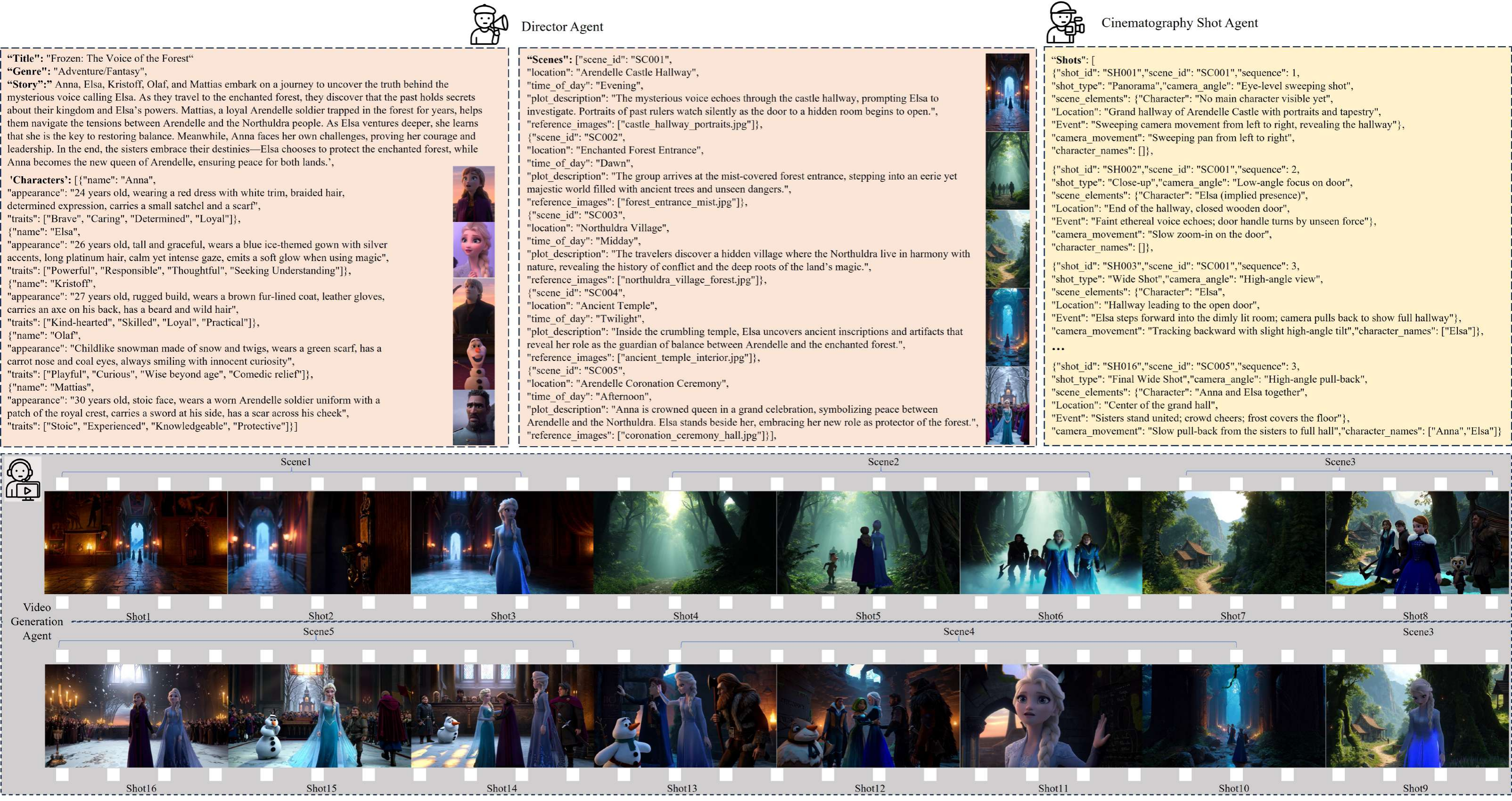} 
    \vspace{-5mm}
\caption{
\textbf{Camera Artist workflow visualization.}
Given a user-provided story outline, Camera Artist decomposes the narrative into structured scene plots and character assets via the \emph{Director Agent}, refines them into coherent shot-level descriptions with explicit cinematic language using the \emph{Cinematography Shot Agent}, and finally renders corresponding visual clips through the \emph{Video Generation Agent}. 
The collaboration among agents enables automated long-form video generation with coherent narrative progression and expressive cinematic shot design.
}
        \vspace{-6mm}
        \label{fig:workflow}
\end{figure*}

\figref{workflow} provides a visual overview of the complete Camera Artist pipeline.
Starting from a textual story outline, the \emph{Director Agent} performs global narrative planning and produces structured assets, including scene-level plots, character attributes, and reference images.
These assets are then consumed by the \emph{Cinematography Shot Agent}, which sequentially generates shot descriptions conditioned on both scene context and previously produced shots, while further enriching each shot with explicit cinematic attributes such as shot size, camera motion, framing, and lighting.
Finally, the \emph{Video Generation Agent} takes the cinematic shot descriptions together with retrieved visual references and synthesizes shot-level video clips, which are temporally concatenated into a long-form narrative video.
This workflow illustrates how Camera Artist operationalizes a film-style production pipeline within a multi-agent system, bridging high-level narrative intent and low-level visual realization.

\begin{figure*}[h]
    \centering
    \includegraphics[width=1\linewidth]{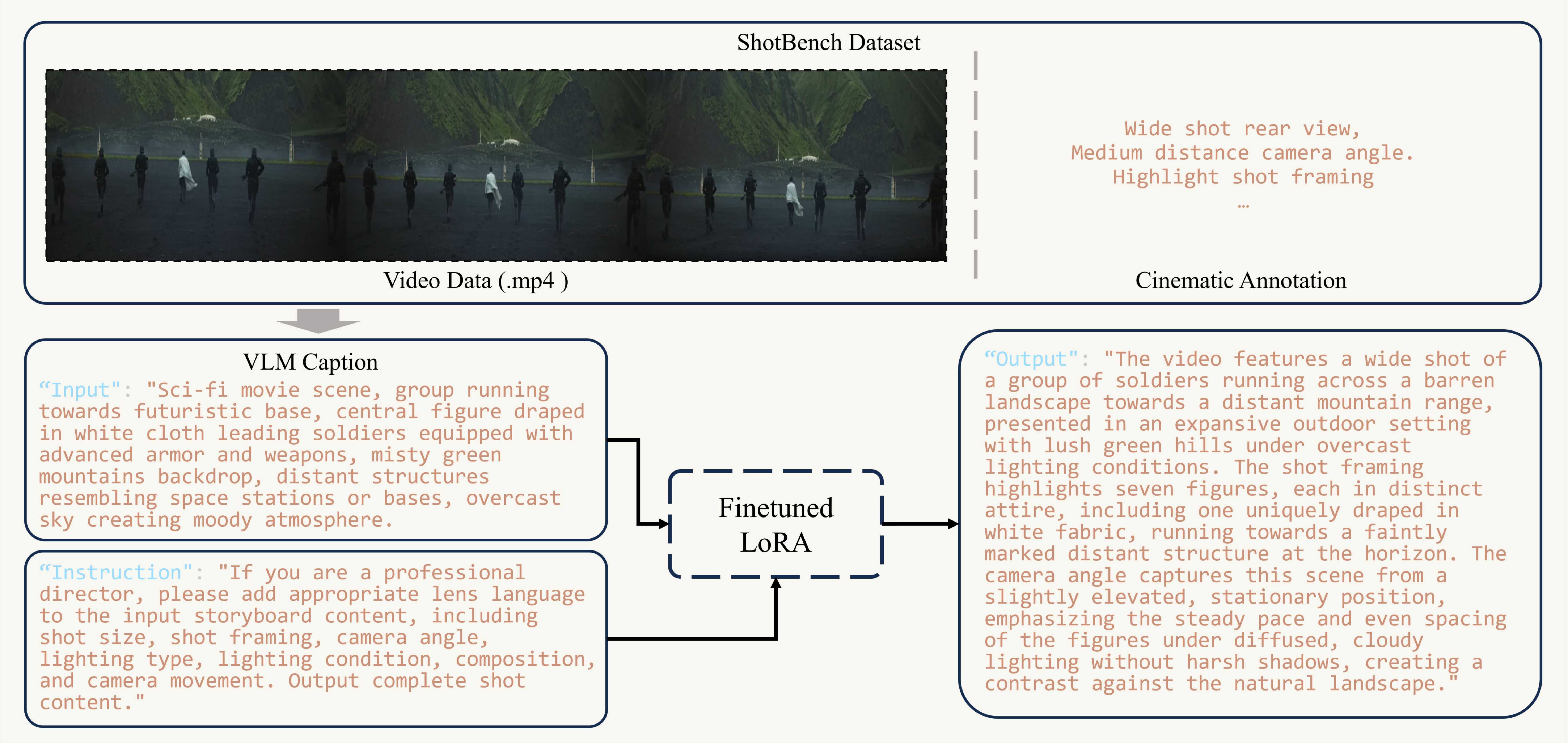} 
        \caption{
            \textbf{An example of pipeline for cinematic language LoRA fine-tuning.}
Ordinary captions are produced by a VLM from raw video, while ShotBench~\cite{liu2025shotbench} provides shot-level cinematic annotations.
A LoRA-tuned LLM learns to transform ordinary captions into cinematic shot descriptions with explicit cinematic language, which are later used for cinematic language injection during inference.
}
\vspace{-6mm}
        \label{fig:lora_demo}
\end{figure*}

\subsection{Cinematic Language LoRA Fine-tuning.}
\figref{lora_demo} illustrates the data construction and fine-tuning process for the Cinematic Language Injection (CLI) module.
We use ShotBench~\cite{liu2025shotbench}, which provides raw video clips together with shot-level cinematic annotations (shot size, angle, framing, motion, lighting).
For each clip, a VLM generates an ordinary caption $x_i$ describing only visible content without cinematic intent.
The target cinematic description $y_i$ is obtained by prompting an LLM to integrate $x_i$ with the corresponding annotation $d_i$, yielding a complete description that explicitly encodes lens language.
We construct 580 paired samples $(x_i, y_i)$ and fine-tune Qwen3-4B using LoRA~\cite{hu2021loralowrankadaptationlarge} (rank 8, scaling factor 32, learning rate $1\times 10^{-4}$, 20 epochs) applied to all linear layers.
The resulting model is used during inference to inject cinematic attributes into recursively generated shot descriptions, which are then fed to the Video Generation Agent.

\subsection{Details of the Chain-of-Thought (CoT) Prompts}
\begin{figure*}[!t]
    \centering
    \includegraphics[width=.97\linewidth]{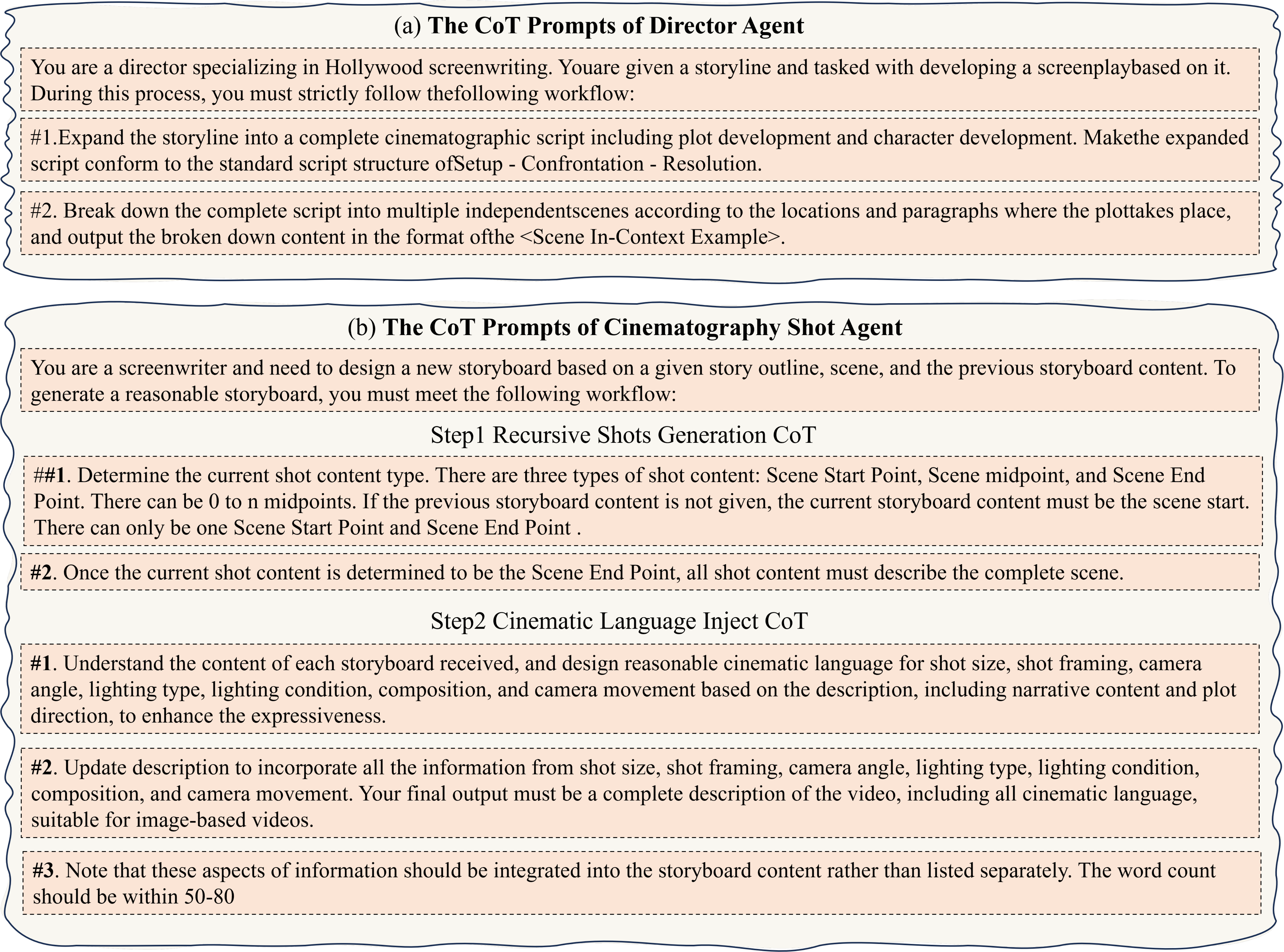} 
    \vspace{-1mm}
        \caption{
            \textbf{The CoT Description of Camera Artist.}
(a) The CoT of Director Agent, which is mainly responsible for the expansion of script content and scene splitting.(b) The CoT of Cinematography Shot Agent, which is mainly responsible for the recursive generation of storyboard content and the introduction of shot language.
        }
        \label{fig:Cotdirector}
    \vspace{-1mm}
\end{figure*}

To clarify how reasoning is performed within our system, we provide diagrammatic illustrations of the CoT~\cite{wei2022chain} prompts used by the Director Agent and Cinematography Shot Agent in \figref{Cotdirector}. 
The Director Agent CoT prompts guides the model to progressively transform a story outline into hierarchical narrative assets by explicitly reasoning through genre, characters, scene objectives, and scene decomposition steps. 
The Cinematography Shot Agent CoT further reasons over previously generated shots and current scene intent, enabling recursive storyboard generation and cinematic decision-making rather than direct, one-step shot output. 
These diagrams illustrate that our agents are not prompted to respond with final answers immediately; instead, they are instructed to “think first and then produce,” making their outputs more structured, coherent, and aligned with real filmmaking logic.

\subsection{Details of Evaluation Details}
\Paragraph{Automatic Evaluation.}
We adopt automatic metrics to objectively assess the quality of generated videos. Following MovieAgent~\cite{wu2025movieagent}, we employ the VBench framework~\cite{huang2023vbench} to evaluate multiple perceptual dimensions, including Subject Consistency, Background Consistency, Motion Smoothness, Dynamic Degree  and Aesthetic Score, using the official VBench~\cite{huang2023vbench} evaluation toolkit and its pretrained video–language backbones\footnote{\url{https://github.com/Vchitect/VBench.git}}. 
To measure semantic faithfulness between the generated videos and the narrative scripts, we further compute CLIP-based text–video similarity using CLIP-T~\cite{radford2021learning}, which extends CLIP with temporal modeling for video understanding. In addition, frame-level semantic alignment is assessed using the CLIP ViT-L/14 image encoder~\cite{radford2021learning}\footnote{\url{https://github.com/openai/CLIP.git}}, providing complementary alignment evaluation between individual frames and textual descriptions. 
Together, these metrics jointly visual quality of the generated videos.

\Paragraph{VLM-Based Evaluation.} 
We employ multiple vision–language models (VLMs) to automatically score generated videos along four dimensions: Script Consistency, Camera-Movement Consistency, Video Quality, and Real-Movie Similarity.
For each metric, we design task-specific prompts that instruct the VLM to analyze the video and output a score from 1 to 5 with a brief justification.

\begin{figure*}[!t]
    \centering
    \includegraphics[width=1.\linewidth]{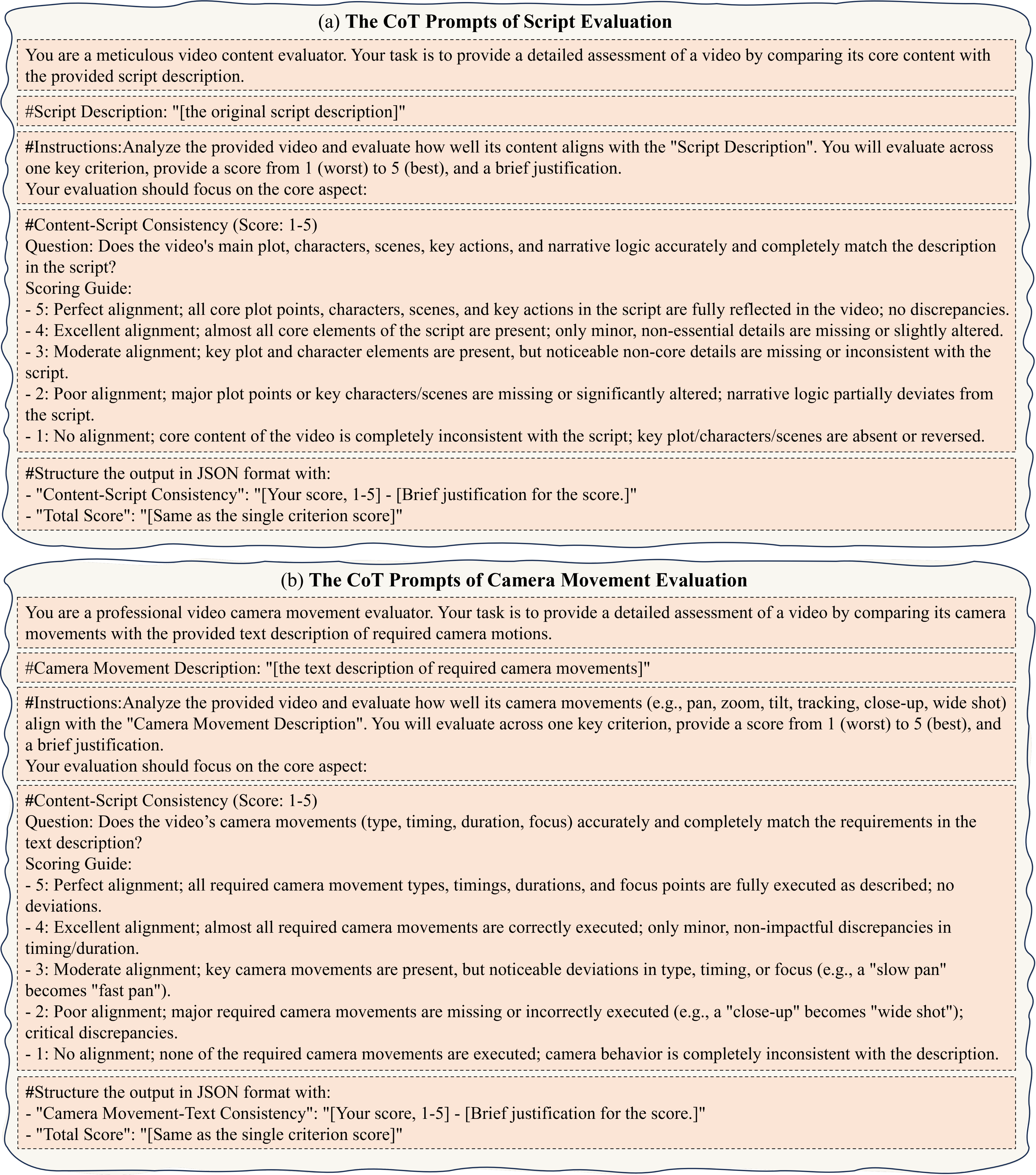} 
        \caption{
            \textbf{The CoT prompting of VLM-based evaluation.}
The CoT is mainly responsible for the recursive generation of storyboard content and the introduction of shot language.
        }
        \vspace{-3mm}
        \label{fig:VLM_evaluation}
\end{figure*}

To reduce redundancy while preserving temporal structure, each video is uniformly sampled into 8–12 keyframes. These keyframes, together with the corresponding textual description (script or camera-motion plan), are provided to the VLM along with one of the four evaluation prompts.
The content of prompts explicitly includes: the evaluator’s role, the evaluation criterion, a scoring rubric from 1 (lowest) to 5 (highest) and required JSON output format (score + explanation), which illustrated as \figref{VLM_evaluation}.

\Paragraph{User Study.} 
The questionnaire follows the same four evaluation dimensions, but the questions are written for human participants rather than for VLM prompts.
For each test case, participants are presented with the input script, anonymized videos produced by different methods.Moreover, Method names are hidden to avoid bias, and the presentation order is randomized.
They then rate each video from 1 (very poor) to 5 (excellent) according to the following questions:
\begin{compactitem}
\item \textbf{Script Consistency:}
How well does the video follow the given script regarding main events, characters, and narrative logic?

\item \textbf{Camera-Movement Consistency:}
How well the camera operations (zoom, pan, tilt, tracking, angle changes, etc.) align with the intended cinematic description and narrative context.

\item \textbf{Video Quality:}
How would you judge the visual quality, clarity, stability, and presence of artifacts?

\item \textbf{Real-Movie Similarity:}
To what extent does the video resemble a real film in cinematography, editing rhythm, color tone, and overall style?
\end{compactitem}

\begin{figure*}[!t]
    \centering
    \includegraphics[width=.98\linewidth]{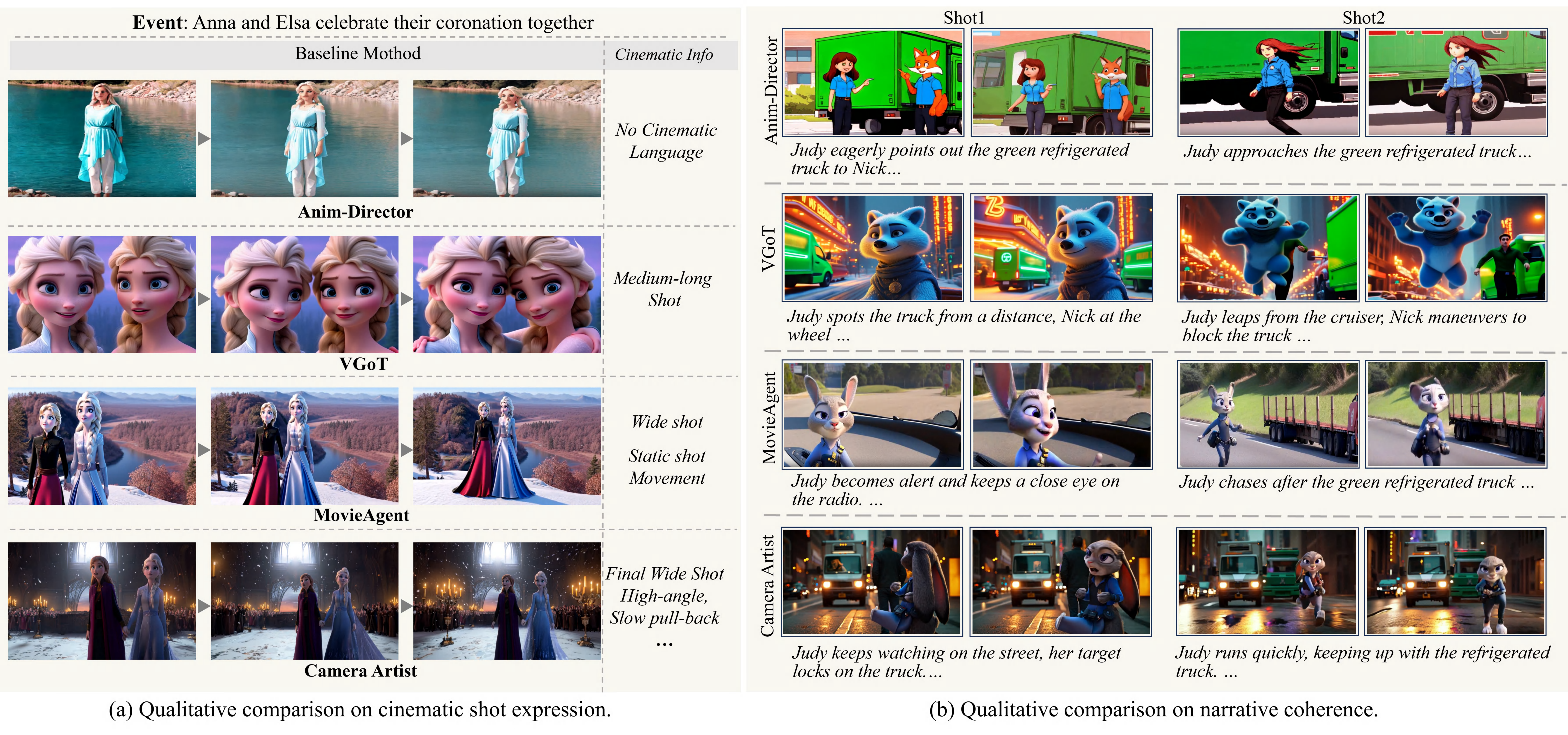} 
        \caption{
            \textbf{Qualitative comparison with baseline methoda.}
(a) Camera Artist generates a final wide shot with high-angle composition and slow pull-back movement, delivering stronger cinematic atmosphere and expressive visual storytelling. (b) Baselines introduce irrelevant characters or exhibit abrupt narrative jumps in two-shot sequences, while Camera Artist maintains both character/scene consistency and coherent event progression.
            %
        }
        \label{fig:sup_qual}
\end{figure*}


\begin{figure*}[!t]
    \centering
    \includegraphics[width=1.\linewidth]{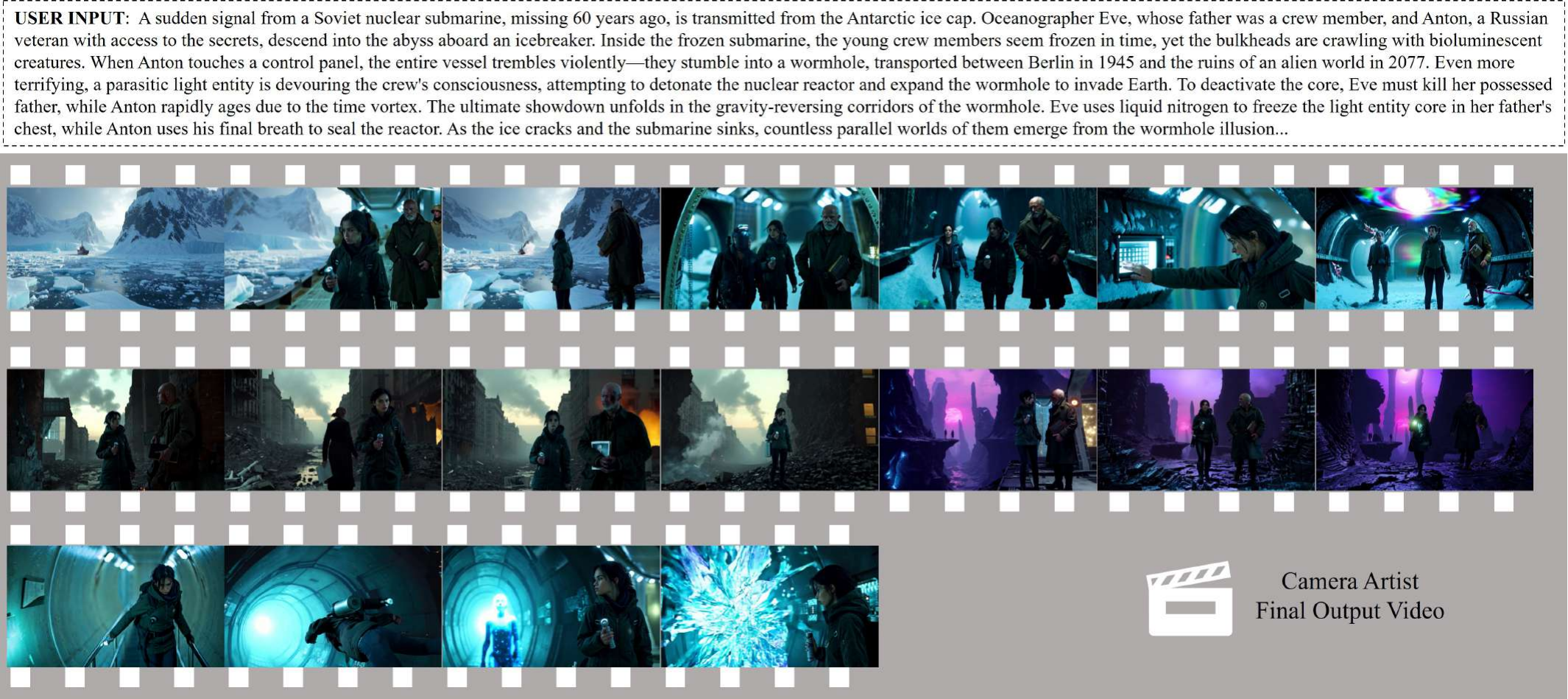} 
        \caption{
            \textbf{Reference-free storytelling video generation.}
Given only a textual story outline (no character reference images), Camera Artist automatically constructs scenes, characters, and shot sequences, producing a long-form narrative video with coherent story progression and cinematic visual expression.
        }
        \label{fig:Reference-free}
\end{figure*}

\begin{figure*}[!t]
    \centering
    \includegraphics[width=1.\linewidth]{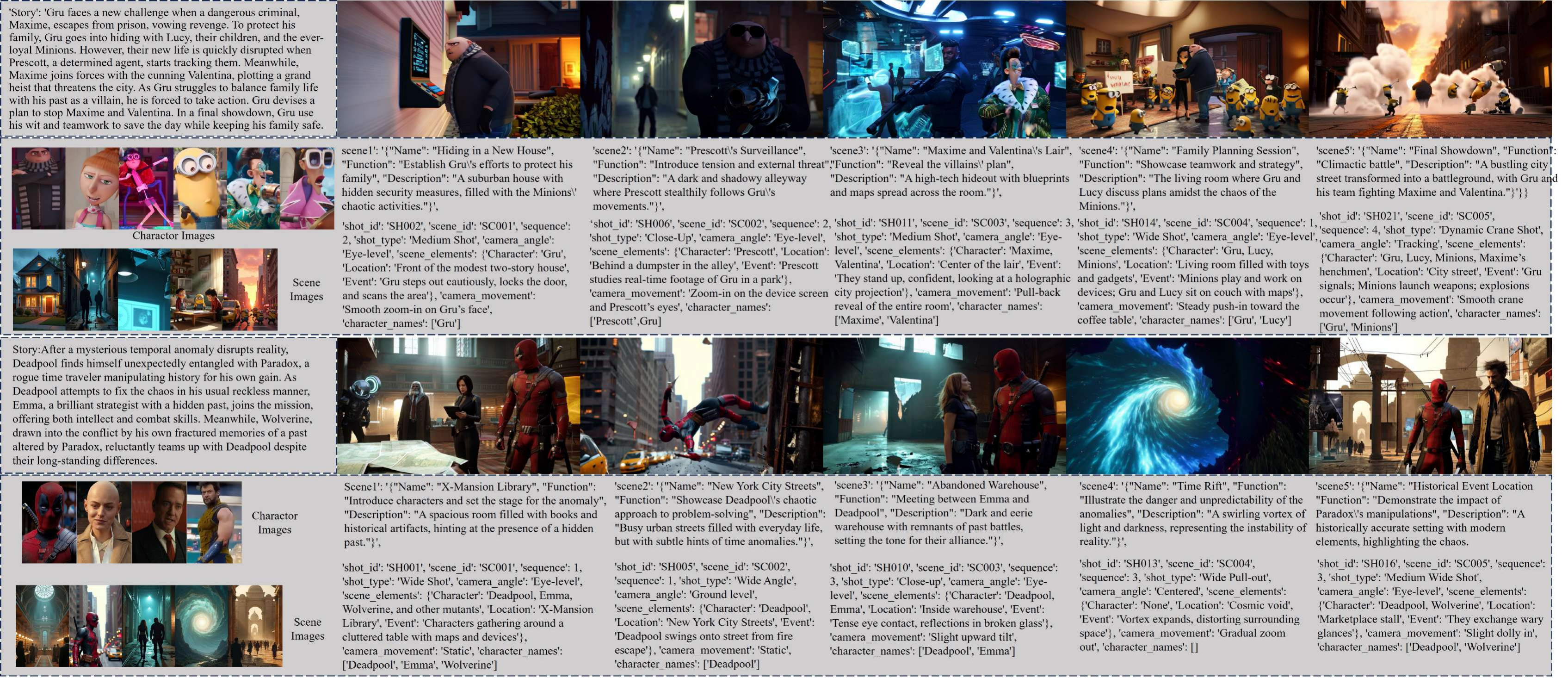} 
        \caption{
            \textbf{Additional qualitative results.}
Scene-level keyframes together with the corresponding  footage are presented, illustrating coherent long-range storytelling, consistent character depiction, and film-style visual expression.
        }
        \label{fig:more_result}
\end{figure*}

\section{Additional Experimental Results}
\label{sec:Additional}

\subsection{Additional Qualitative Comparison.}
\figref{sup_qual} (a) presents an additional qualitative comparison on the event ``Anna and Elsa celebrate their coronation together.'' 
Baseline systems are able to produce visually plausible video frames, yet their cinematic expressiveness remains limited. 
Anim-Director~\cite{10.1145/3680528.3687688} mainly outputs static framings without explicit lens design. 
VGoT~\cite{zheng2024videogen} produces medium--long shots but lacks purposeful camera control. 
MovieAgent~\cite{wu2025movieagent} is able to generate wide shots, yet the camera remains largely static, resulting in weak visual dynamics. 
In contrast, Camera Artist adopts a deliberately designed final wide shot with high-angle composition and slow pull-back camera movement, which not only highlights ceremonial atmosphere but also strengthens emotional emphasis and film-like presentation. 
This example further illustrates the advantage of our framework in generating shots with richer cinematic language rather than merely depicting scene content.

We also provided an additional result of inter-shot narrative coherence in \figref{sup_qual} (b).
In this example, two consecutive shots are intended to jointly depict the event of Judy independently tracking the refrigerated truck.
Anim-Director~\cite{10.1145/3680528.3687688} and VGoT~\cite{zheng2024videogen} incorrectly introduce an extra character (Nick), leading to semantic drift and identity inconsistency. MovieAgent~\cite{wu2025movieagent} preserves character identity, but its narrative jumps abruptly from waiting for radio messages to chasing the truck, breaking event continuity.
In contrast, Camera Artist depicts a coherent progression—Judy discovers the truck and then closely follows it—while maintaining stable character and scene consistency across shots.

\subsection{Storytelling without character reference images.}
Benefiting from the powerful generative capability of modern T2I models and multi-reference I2V tools, our framework is not limited to cases where character reference images are provided. 
Camera Artist can also operate in a \emph{reference-free} setting, where only a textual story outline is given and both characters and scenes are automatically synthesized during generation.
This enables fully automated long-form storytelling video generation from pure text, while still preserving narrative coherence and expressive cinematic presentation.
\figref{Reference-free} shows an example of a long narrative generated solely from a textual story description without any character reference images.

\subsection{More Qualitative Results}
To further demonstrate the effectiveness and generality of Camera Artist, we present additional qualitative results.
For each story, we visualize scene-level keyframes that summarize the visual progression within individual scenes and footage sequences covering the entire narrative as shown in \figref{more_result}. The scene keyframes highlight how our framework maintains character identity, spatial continuity, and cinematic style across scenes, while the complete footage illustrates long-range narrative coherence, smooth shot transitions, and consistent visual storytelling across complex multi-scene plots.


\end{document}